\documentclass{article}

\PassOptionsToPackage{numbers,sort&compress}{natbib}
\usepackage[dblblindworkshop,final]{neurips_2026}

\usepackage[utf8]{inputenc}
\usepackage[T1]{fontenc}
\usepackage{booktabs}
\usepackage{graphicx}
\usepackage{amsmath}
\usepackage{microtype}
\usepackage[hidelinks]{hyperref}
\usepackage{caption}
\title{4MT-VLM: How Coarse Is a VLMs Cognitive Map?}

\author{Markus Frey \\
  Lamarr Institute for Machine Learning and Artificial Intelligence\\
  Fraunhofer IAIS, University of Bonn\\
  \texttt{markus.frey@iais.fraunhofer.de}
}

\begin{document}
\maketitle

\begin{abstract}
An agent that moves must recognise a place from a viewpoint it has never seen. We introduce 4MT-VLM, a dataset of procedurally generated landscapes, each rendered across five stimulus modes that remove appearance cues while holding layout fixed: shape and colour, shape only, colour only, bare terrain peaks with no objects, and a valley viewpoint that puts the peaks on the horizon. The last condition is commonly used in clinics to probe hippocampal function in human patients. We test this benchmark across sixteen different open and closed-source models and report 4AFC performance, a measure which is also used to grade human participants. We observe that models identify a place from the studied viewpoint but lose it once the camera moves, dropping below the 25\% chance level at 135° where a human observer scores 85\%. Frontier models (Gemini 3.8 Flash, GPT-5.6) answer only 39\% and 31\% of rotated trials correctly, recovering to 85\% and 55\% only when distractors are moved more than 30 meters apart. Our benchmark demonstrates that while current VLMs possess rudimentary cognitive maps, their spatial resolution remains fundamentally too coarse to maintain a stable, 3D understanding of the world once the viewpoint changes.
\end{abstract}

\section{Introduction}

An agent that moves must recognise a place from a viewpoint it has never seen. Doing so requires an allocentric representation, i.e. a cognitive map of where things are that does not depend on where the observer stands \citep{tolman1948cognitive,okeefe1978hippocampus}. As vision--language models (VLMs) are increasingly deployed as the visual reasoning engines for embodied agents, they are expected to maintain a stable, 3D understanding of the world as they navigate and their camera view changes. 

How well they actually do this is hard to read off current spatial benchmarks. Most multimodal evaluations score perception, language, and geometry together, returning a single accuracy number. A low score indicates that something is missing without identifying what, and a high score often means the model found a visual shortcut, such as matching textures or background colours. Furthermore, difficulty in these benchmarks is arbitrary. Because embodied agents operate in physical space, their spatial reasoning failures should be measurable in physical units, not just percentage points on a static dataset.

To build a more diagnostic evaluation, we adapt The Four Mountains Test (4MT) \citep{hartley2007hippocampus}, which was designed specifically to isolate allocentric spatial memory. A participant studies a computer-generated landscape, then must identify it among four candidates rendered from a new viewpoint, with all colours and textures resampled. Scores fall with hippocampal damage \citep{hartley2007hippocampus,bird2010topographical} and in pre-dementia Alzheimer's disease \citep{chan2016four,wood2016allocentric}, which attacks the exact brain regions responsible for allocentric navigation. 

For benchmarking allocentric perception, we build a novel test with two primary constraints: 
\begin{enumerate}\itemsep2pt
\item \textbf{Viewpoint change} $\Delta \in \{0^\circ, 45^\circ, 90^\circ, 135^\circ, 180^\circ\}$. Appearance is resampled at every $\Delta$, so $\Delta=0$ is not an image match but a check that the place can be identified at all. We call accuracy at $\Delta=0$ the \emph{appearance gate}, and treat it as the denominator for any claim about viewpoint invariance.
\item \textbf{Distractor similarity, in metres.} For every pair of scenes we compute the rotation-optimal layout distance $D(i,j)=\min_\phi \frac{1}{K}\sum_k \lVert R(\phi)p_{ik}-p_{jk}\rVert$, which is what is left after the best rigid rotation, and draw distractors from a chosen percentile band of that distribution.
\end{enumerate}

Crossing these two axes separates three distinct failures that standard benchmarks run together: failing to identify the scene visually, failing to compensate for camera rotation, and failing to distinguish physical layouts that are too close together. Our contributions are:

\begin{itemize}\itemsep2pt
\item \textbf{4MT-VLM:} A diagnostic benchmark for embodied AI, adapted from a clinical test of hippocampal function. It contains 500 trials over 100 procedurally generated landscapes, systematically stripping away appearance cues to isolate layout across graded viewpoint shifts.
\item \textbf{Evaluation in physical units.} We generate metric ``hard negatives'' by drawing distractors from percentile bands of layout distance in metres. Because we redraw only the distractors while leaving the target and azimuths identical, differences in performance represent a measurable limit on the model's spatial resolution.
\item \textbf{A physical diagnosis of scaling failure.} We show that scaling open-weight models from 1B to 235B parameters drastically improves scene identification (the appearance gate) but does nothing for rotational invariance, leaving performance at or below chance. Distance is the only manipulation that recovers performance: Gemini 3.8 Flash jumps from 39\% to 85\% accuracy only when distractors are moved 31\,m apart. The information is there, but its resolution is fundamentally too coarse for precise allocentric navigation.
\end{itemize}

\section{Related work}

\paragraph{Spatial memory in humans.}
The 4MT turns the cognitive map account of hippocampal function \citep{okeefe1978hippocampus} into a four-alternative forced choice, and its diagnostic value comes from the viewpoint shift specifically
\citep{hartley2007hippocampus}. Separately, \citet{shepard1971mental} showed that the time to match two objects across a rotation grows linearly with angular disparity, which indicates a transformation applied to a representation rather than a lookup of a stored view. Both paradigms fix the scene, move the vantage point, and measure what that costs. \citet{frey2023probing} take the same task to artificial networks, training them on a scene-perception problem while reading out what the learned representations encode.

\paragraph{Spatial benchmarks for vision--language models.}
Recent evaluations keep finding the same split between egocentric and allocentric questions. \citet{fu2024blink} turn fourteen classic vision tasks into multiple choice and report frontier accuracy near 50\% where people score above 95\%. \citet{yang2024thinking} measure spatial memory from video and find that chain-of-thought, self-consistency and tree-of-thoughts all fail to help. \citet{li2025viewspatial} find competent reasoning in the camera frame and failure in another entity's frame. \citet{zhang2025spinbench} isolate perspective taking and rotation across 43 models and report a strong egocentric bias against 91\% human accuracy, and \citet{zhang2026multiview} find that models handle 2D relations within one image but not the combination of several views into one global frame. These benchmarks establish that the capability is weak. Difficulty in them comes from the choice of question, not from a measured property of the individual item, so they can report that a model fails without locating the point where it starts to.

\paragraph{Building allocentric representations.}
A parallel line of work supplies the missing representation instead of measuring it.  \citet{gu2026spacemind} build a voxelised cognitive map from video and fuse it back into the visual features. \citet{ruan2026world2mind} build a top-down landmark tree using reconstruction and segmentation tools, and report that a text-only model reading that tree comes close to multimodal performance. Both assume the representation is missing and add it from outside. We ask instead what the representation already inside the model supports, and where it runs out of resolution.

\section{Methods}

\begin{figure}[t]
\centering
\includegraphics[width=\textwidth]{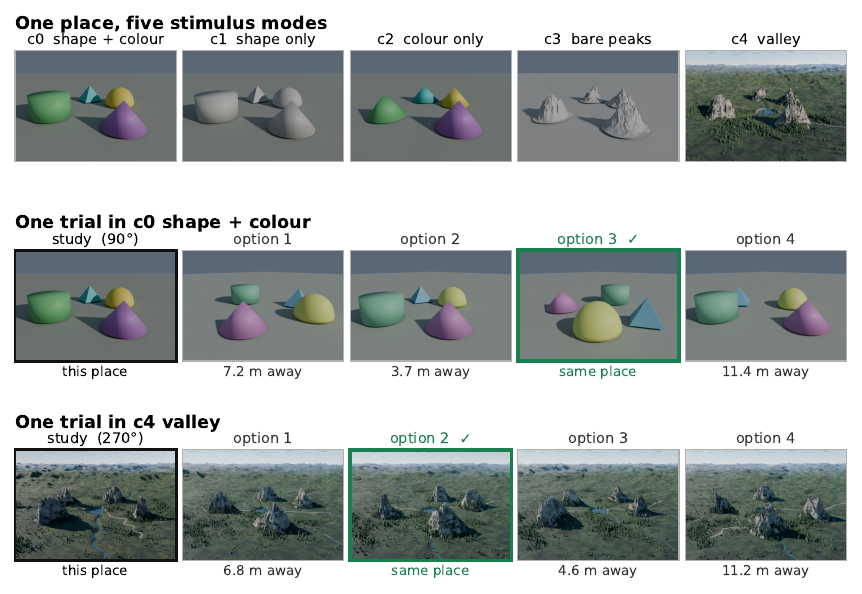}
\caption{\textbf{Top:} one place in the five stimulus modes, at one viewpoint.
Colour is removed, then shape, then the objects themselves, and c4 moves the
camera into the valley so the peaks sit on the horizon. Layout is identical
across the five. \textbf{Middle and bottom:} one trial in c0 and one in c4. The
study view, then four candidates rendered $135^\circ$ away with appearance
resampled, labelled with their layout distance to the target.}
\label{fig:stimuli}
\end{figure}

\paragraph{Stimuli.}
We rendered 100 procedurally generated landscapes of four peaks each in Blender, from eight azimuths under two appearance samples. Five stimulus \emph{modes} strip appearance cues away while holding layout fixed: shape and colour (c0), shape only (c1), colour only (c2), bare terrain peaks with no objects (c3), and a valley viewpoint with the peaks on the horizon (c4) (see Figure~\ref{fig:stimuli}). All modes draw from a shared inventory of objects, so that no mode carries landmark identity that another lacks.

\paragraph{Task.}
Each trial shows a study image and four candidates rendered at a common test azimuth, one of which is the study scene. Study and test images always use different appearance samples, at every $\Delta$. The full set is 500 trials, balanced over 5 modes $\times$ 5 values of $\Delta$ $\times$ 20 items, with the correct answer spread evenly over the four response positions.

\paragraph{Distractor sets.}
Distractors come from a percentile band of $D(i,j)$, computed separately within each mode because the modes differ in absolute scale. The main set uses the 0--10th percentile, which gives a median nearest-distractor distance of 6.8\,m. Two matched sets redraw only the distractors, from the 40--60th percentile (31.4\,m) and the 90--100th (42.8\,m).

\paragraph{Landmark count.}
The distractor sets vary how far the alternatives sit from the target while holding the scenes fixed. A second manipulation varies the number of objects in the scenes. We rendered three further banks of 100 landscapes containing one, two and six landmarks, identical to the main bank in every render setting (eight azimuths, $640\times 440$, 24 samples, same seed), and built a matched four-landmark bank from the main scenes so that all four sizes are constructed the same way. These use the object mode (c0) only, because every landmark must be unique in both shape and colour and the inventory holds six of each.

\paragraph{Procedure.}
Every observer answered the same 100-trial subset, stratified over mode and $\Delta$. Models got a chain-of-thought instruction, a token budget of 8000 and, where the provider exposes it, medium reasoning effort. Every instruction is reproduced verbatim in the appendix.

\paragraph{Analysis.}
Chance level for the four-alternative forced choice (4AFC) task is 25\%. Confidence intervals are Wilson intervals, paired comparisons on identical trials use the McNemar test, and distributional comparisons use permutation tests with $2\times10^4$ resamples. We checked that no observer's answer distribution alone gives an advantage: the accuracy expected from each observer's own answer frequencies, ignoring the images, is between 19\% and 27\% for all seventeen observers.

\section{Results}

We report two accuracies throughout. The appearance gate is the accuracy at $\Delta=0$, where the place is shown from the studied viewpoint under a fresh appearance sample, e.g. the scene did not change but the position of the sun did. The rotated accuracy is the performance of the model for all other images where the viewpoint change is $\Delta\ge45^\circ$. 

\begin{table}[t]
\centering
\small
\setlength{\tabcolsep}{6pt}
\begin{tabular}{@{}lrrrrr@{}}
\toprule
\textbf{Observer} & \textbf{Params} & \textbf{Accuracy} & \textbf{95\% CI} & \textbf{$\Delta\!=\!0$ gate} & \textbf{$\Delta\!\geq\!45$} \\
\midrule
Human & -- & \textbf{86\%} & [78\%, 91\%] & \textbf{100\%} & \textbf{82\%} \\
\midrule
InternVL3.5-4B & 4B & 17\% & [11\%, 26\%] & 35\% & 12\% \\
InternVL3.5-2B & 2B & 21\% & [14\%, 30\%] & 25\% & 20\% \\
InternVL3.5-1B & 1B & 23\% & [16\%, 32\%] & 35\% & 20\% \\
InternVL3.5-38B & 38B & 28\% & [20\%, 37\%] & 65\% & 19\% \\
InternVL3.5-8B & 8B & 28\% & [20\%, 37\%] & 55\% & 21\% \\
InternVL3.5-30B-A3B & 30B & 29\% & [21\%, 39\%] & 55\% & 22\% \\
Qwen2.5-VL-32B & 32B & 29\% & [21\%, 39\%] & 75\% & 18\% \\
Qwen2.5-VL-3B & 3B & 29\% & [21\%, 39\%] & 30\% & 29\% \\
Qwen3-VL-235B (t) & 235B & 30\% & [22\%, 40\%] & 75\% & 19\% \\
InternVL3.5-14B & 14B & 31\% & [23\%, 41\%] & 50\% & 26\% \\
Qwen2.5-VL-72B & 72B & 31\% & [23\%, 41\%] & 75\% & 20\% \\
Qwen3-VL-235B (i) & 235B & 33\% & [25\%, 43\%] & 90\% & 19\% \\
InternVL3-38B & 38B & 34\% & [25\%, 44\%] & 85\% & 21\% \\
Qwen2.5-VL-7B & 7B & 36\% & [27\%, 46\%] & 55\% & 31\% \\
Gemini 3.8 Flash & -- & 44\% & [35\%, 54\%] & 65\% & \textbf{39\%} \\
GPT-5.6 Luna & -- & \textbf{45\%} & [36\%, 55\%] & \textbf{100\%} & 31\% \\
\bottomrule
\end{tabular}
\caption{Four-alternative forced choice with hard foils (100 trials, chance 25\%: 20 unrotated and 80 rotated). The $\Delta=0$ column is the appearance gate. Qwen3-VL-235B (i) is the \texttt{instruct} variant, (t) the \texttt{thinking} one. Rows are ordered by overall accuracy. Bold marks the human row and the best model in each column. Accuracy by mode is reported in Appendix Table~\ref{tab:modes}.}
\label{tab:main}
\end{table}

\paragraph{Models identify the scene and lose it under rotation.}
Table~\ref{tab:main} reports the two accuracies separately. GPT-5.6 Luna answers 100\% of unrotated trials and 31\% of rotated trials. Qwen3-VL-235B answers 90\% and 19\%, where a human observer answers 100\% and 82\%. On the identical 100 trials, paired exact McNemar puts every model below the human baseline ($p=1.3\times10^{-10}$ for Gemini 3.8 Flash, $1.3\times10^{-8}$ for GPT-5.6 Luna, $2.3\times10^{-14}$ for Qwen3-VL-235B), and the gap holds within rotated trials alone ($p\le1.3\times10^{-8}$ for all sixteen). The task is passable, and the viewpoint change is the part the models fail.

\begin{figure}[t]
\centering
\includegraphics[width=\textwidth]{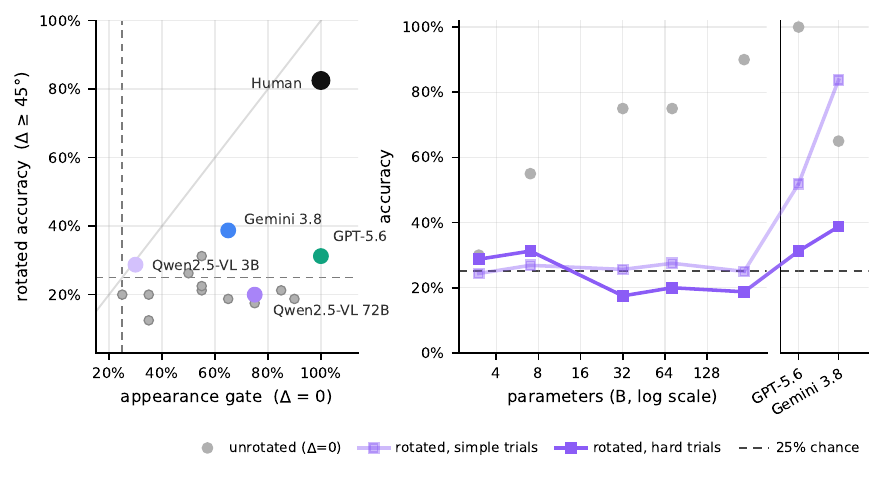}
\caption{\textbf{Left:} unrotated against rotated accuracy. An observer working from appearance sits in the lower right; one that cannot identify the scene at all sits in the lower left. Grey points are the models tested but without labels in this figure. \textbf{Right:} the same two quantities against parameter count, on the identical trials, for Qwen2.5-VL at 3B, 7B, 32B and 72B and Qwen3-VL-235B in \texttt{instruct} mode. Grey is unrotated accuracy, which rises with scale while rotated accuracy does not, on hard trials or on simple ones. InternVL3.5 runs the same ladder and lands on top of the Qwen line and is reported in Table~\ref{tab:main}.}
\label{fig:gate}
\end{figure}

\paragraph{Scale buys scene identification, not viewpoint invariance.}
Across Qwen2.5-VL at 3B, 7B, 32B and 72B, with prompt, stimuli and decoding fixed, unrotated accuracy rises $30\to55\to75\to75\%$ while rotated accuracy falls $29\to31\to18\to20\%$ and crosses below chance
(Figure~\ref{fig:gate}). To rule out a floor effect from distractor difficulty, we repeated the series with distractors $4.6$ to $6.3\times$ further away. Rotated accuracy stays flat at $24\to27\to26\to28\%$. On the same trials the identical change takes Gemini 3.8 Flash from 39\% to 85\% and GPT-5.6 Luna from 31\% to 55\%. Another generation at three times the size does not change the picture. Qwen3-VL-235B-A22B identifies 90\% of unrotated scenes and answers 19\% of rotated ones; widening the distractors takes its gate to 100\% and its rotated accuracy to 25\%, which is chance. More inference-time computation does not help either: the \texttt{thinking} variant reaches 30\% overall against 33\% for \texttt{instruct}, with the same 19\% rotated.

To test the scaling beyond the Qwen family, we ran InternVL3.5 at 1B, 2B, 4B, 8B, 14B and 38B on the same trials with the same prompt and token ceiling. Unrotated accuracy rises from 25\% to 65\% between 2B and 38B, while rotated accuracy reads $20, 20, 12, 21, 26, 19\%$ across the whole $38\times$ range and never clears the 25\% chance level. Its mixture-of-experts member (30B total, 3B active) behaves like its dense neighbours, at 55\% and 22\%. Across all fourteen open-weight models in the set, spanning 1B to 235B parameters, no rotated accuracy passes 31\% (chance at 25\%).

\begin{table}[t]
\centering
\small
\setlength{\tabcolsep}{6pt}
\begin{tabular}{@{}lccccc@{}}
\toprule
 & \includegraphics[width=0.122\textwidth]{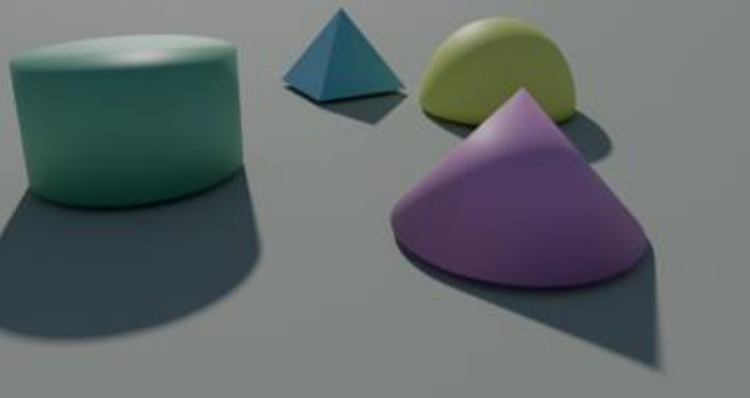} & \includegraphics[width=0.122\textwidth]{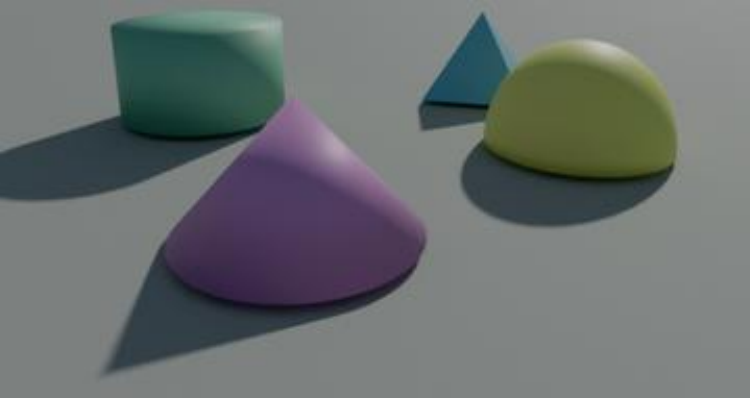} & \includegraphics[width=0.122\textwidth]{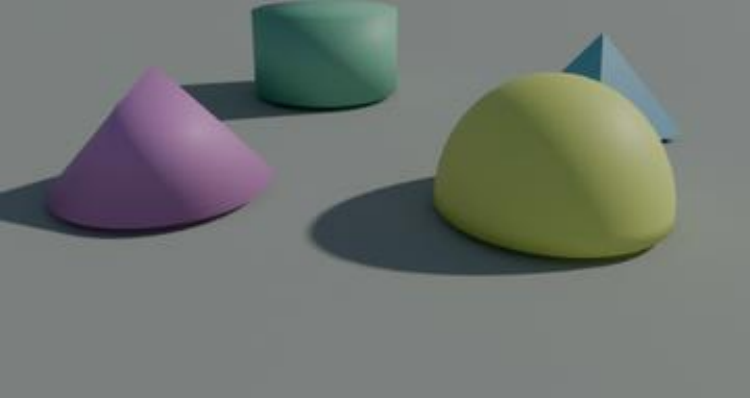} & \includegraphics[width=0.122\textwidth]{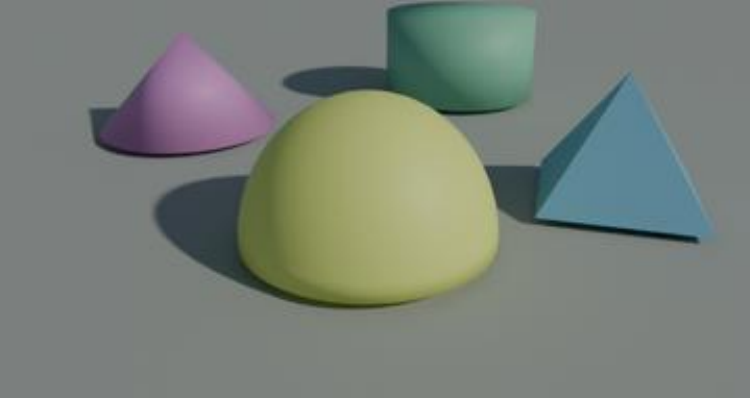} & \includegraphics[width=0.122\textwidth]{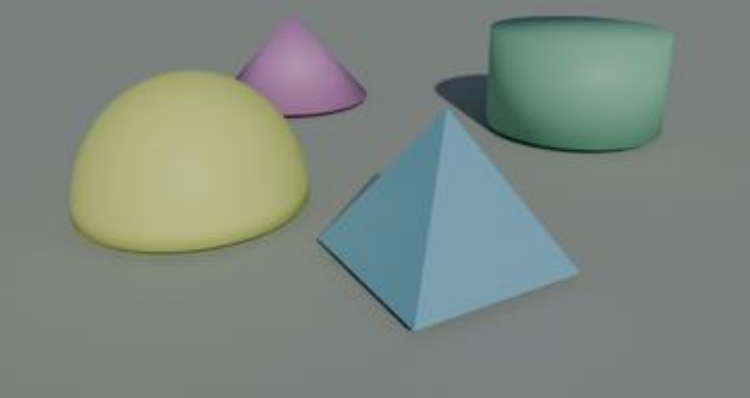} \\[1pt]
\textbf{Observer} & \textbf{$\Delta\!=\!0^\circ$} & \textbf{$\Delta\!=\!45^\circ$} & \textbf{$\Delta\!=\!90^\circ$} & \textbf{$\Delta\!=\!135^\circ$} & \textbf{$\Delta\!=\!180^\circ$} \\
\midrule
Human & 100\% & 90\% & 80\% & 85\% & 75\% \\
\midrule
InternVL3.5-4B & 35\% & 20\% & 20\% & 5\% & 5\% \\
InternVL3.5-2B & 25\% & \textbf{10\%} & 25\% & 20\% & 25\% \\
InternVL3.5-1B & 35\% & 15\% & 30\% & 25\% & \textbf{10\%} \\
InternVL3.5-38B & 65\% & 15\% & 15\% & 25\% & 20\% \\
InternVL3.5-8B & 55\% & 40\% & 20\% & \textbf{5\%} & 20\% \\
InternVL3.5-30B-A3B & 55\% & 35\% & 15\% & 25\% & 15\% \\
Qwen2.5-VL-32B & 75\% & 30\% & 15\% & \textbf{0\%} & 25\% \\
Qwen2.5-VL-3B & 30\% & 50\% & \textbf{15\%} & 25\% & 25\% \\
Qwen3-VL-235B (t) & 75\% & 30\% & 25\% & 10\% & 10\% \\
InternVL3.5-14B & 50\% & 30\% & 25\% & \textbf{20\%} & 30\% \\
Qwen2.5-VL-72B & 75\% & 40\% & 15\% & \textbf{10\%} & 15\% \\
Qwen3-VL-235B (i) & 90\% & 40\% & 15\% & \textbf{5\%} & 15\% \\
InternVL3-38B & 85\% & 40\% & \textbf{5\%} & 15\% & 25\% \\
Qwen2.5-VL-7B & 55\% & 45\% & \textbf{15\%} & 25\% & 40\% \\
Gemini 3.8 Flash & 65\% & 45\% & 50\% & \textbf{15\%} & 45\% \\
GPT-5.6 Luna & 100\% & 35\% & 35\% & \textbf{10\%} & 45\% \\
\midrule
\textbf{All models pooled} & \textbf{61\%} & \textbf{32\%} & \textbf{21\%} & \textbf{15\%} & \textbf{23\%} \\
\bottomrule
\end{tabular}
\caption{Accuracy by viewpoint change. The images above the columns are one c0 scene at each viewpoint change, from the same study view. Pooled over the models the floor is $\Delta=135^\circ$, below chance, and accuracy recovers at a half turn. Bold marks each model's own worst viewpoint; rows that tie for worst are left unmarked.}
\label{tab:delta}
\end{table}

\paragraph{The worst viewpoint change across model is $135^\circ$.}
Pooled over the sixteen models, accuracy at $135^\circ$ is 48/320 = 15.0\%, clearly below chance (one-sided binomial $p=9\times10^{-6}$), and recovers to 23\% at $180^\circ$ (Table~\ref{tab:delta}). Taking each model's own worst viewpoint, seven bottom out at $135^\circ$, three at $90^\circ$, one at $45^\circ$ and one at $180^\circ$, and four tie across two viewpoints. A $180^\circ$ turn is the largest change and the condition that should be hardest if accuracy simply decayed with angle. It suggests the models are applying an image heuristic, such as matching against a reflection of the original scene, which partially succeeds at a direct reversal but misleads the model at intermediate angles.

\paragraph{Distractor separation separates frontier from open models.}
Raising the median nearest-distractor distance from 6.8\,m to 31.4\,m on identical trials takes Gemini 3.8 Flash from 39\% to 85\% (exact McNemar $p=1.5\times10^{-10}$) and GPT-5.6 Luna from 31\% to 55\% ($p=2\times10^{-3}$;
Figure~\ref{fig:gate}). The same change moves no open model significantly at either wider separation. From 6.8\,m to 31.4\,m, Qwen2.5-VL at 3B, 7B, 32B and 72B moves by $-4$, $-6$, $+10$ and $+4$ points ($p=0.65$, $0.36$, $0.13$,
$0.63$), and InternVL3-38B by $-3$ ($p=0.66$). From 6.8\,m to 42.8\,m the largest shift in the open set is $+11$ points, for Qwen2.5-VL-72B ($p=0.08$).

\begin{figure}[t]
\centering
\includegraphics[width=\textwidth]{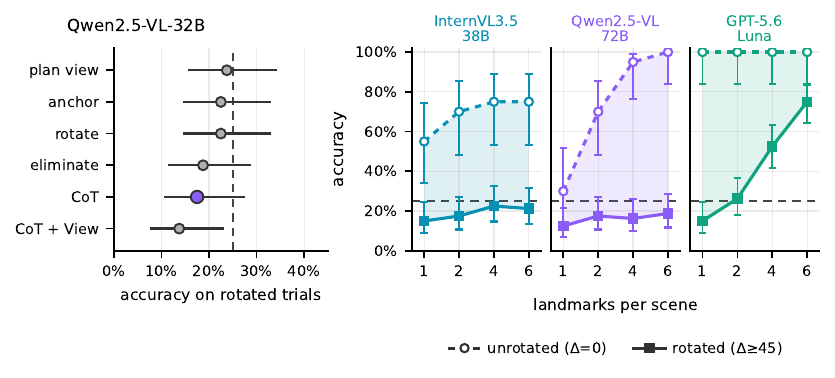}
\caption{\textbf{Left:} Qwen2.5-VL-32B under six instruction styles, rotated
trials only, with the instruction used everywhere else highlighted. The styles
are, from the top, "imagine a plan view", "pick an anchor landmark", "rotate the
scene mentally", "eliminate alternatives", plain chain of thought, and chain of
thought with the viewpoint change asserted. None reaches chance.
\textbf{Right:} accuracy against the number of landmarks in a scene, one panel
per model. The shaded band is the gap between identifying a place and
recognising it after a viewpoint change. Adding landmarks raises the unrotated
line in all three models and closes the gap in only one. Landmark count and
layout separation cannot be varied independently: the median distractor sits at
0.9, 7.6, 23.2 and 28.6\,m for one, two, four and six landmarks.}
\label{fig:landmarks}
\end{figure}

\paragraph{Making the task easier raises the gate and leaves rotation alone.}
Varying the number of landmarks gives a second route to the same conclusion, with different scenes rather than different distractors (Figure~\ref{fig:landmarks}). Qwen2.5-VL-72B goes from identifying 30\% of one-landmark scenes to 100\% of six-landmark scenes, and its rotated accuracy over the same range moves from 12\% to 19\%.
InternVL3.5-38B behaves the same way, 55\% to 75\% unrotated against 15\% to 21\% rotated. GPT-5.6 Luna identifies every scene at every size, while its rotated accuracy rises from 15\% to 75\%.

\paragraph{No instruction style recovers the missing operation.}
Qwen2.5-VL-32B answered the same 80 rotated trials under six instructions, with stimuli and decoding held fixed (see Appendix~\ref{app:prompts} for exact prompts). \texttt{cot}, used for every other run in the paper, says the matching option may be viewed from the study direction or from a different one, while \texttt{cot + view} states that it is viewed from a different viewpoint. The other four keep that framing and add an explicit strategy: estimate the rotation angle and mentally rotate the peaks, checking whether their clockwise order is preserved (\texttt{mental\_rotation}); take the most distinctive peak as an origin and trace the bearings and distances of the other three from it
(\texttt{anchor}); imagine the layout from directly above and work out where the camera would stand on that map (\texttt{birdseye}); or look for a geometric contradiction in each candidate and eliminate it (\texttt{elimination}). Accuracy is 23.8\% for \texttt{birdseye}, 22.5\% for \texttt{mental\_rotation} and \texttt{anchor}, 18.8\% for \texttt{elimination}, 17.5\% for
\texttt{cot} and 13.8\% for \texttt{cot + view}. None reaches chance (see Figure~\ref{fig:landmarks}).

\paragraph{Models fail on the same trials, and their errors ignore layout.}
If the models were failing for unrelated reasons they would fail on unrelated trials, which they do not. Cohen's $\kappa$ on trial-level correctness, over all 100 trials, averages 0.18 across the 120 model pairs and 0.05 across the 16 human--model pairs (permutation $p=5\times10^{-4}$; Appendix
Figure~\ref{fig:agree}), so the models resemble each other more than any of them resembles a human.  Where the errors land is also informative. Each wrong answer picks one of three foils, and we rank those by layout distance to the target: nearest, middle, farthest. An observer with a graded sense of layout should confuse a place with its nearest neighbour most often. Pooled over 1092 model errors the split is 30/37/33, against the 33/33/33 expected from choices made without regard to geometry, and no individual model departs from a third (Appendix Figure~\ref{fig:agree}), while a human picks the \textit{nearest} foil 10 out of 14 times.

\section{Discussion}

Every model we tested identifies a place from the studied viewpoint and then loses it once the camera moves. Perception is not the bottleneck as we show that with six landmarks, Qwen2.5-VL-72B identifies
100\% of unrotated scenes and answers 19\% of the rotated ones, and InternVL3.5-38B behaves the same way. A model that resolves every object in the scene well enough to recognise it from the studied viewpoint has demonstrably seen what it needs to see. Scene complexity is not the bottleneck either: the landmark-count banks vary the scenes instead of the distractors and produce the same split, with rotated accuracy tracking how far apart the layouts are rather than how much is in them.

The trials are answerable. A human observer, given the same 100 items and the distractors at their closest spacing, answers 82\% of the rotated ones. The information needed is present in the images, and enough of it survives the viewpoint change for at least one observer to use.

Scale does not supply the missing operation. Between 3B and 72B, Qwen2.5-VL gains 45 points on the appearance gate and loses 9 on rotated trials, and InternVL3.5 repeats the pattern from 1B to 38B without ever clearing chance. Neither does inference-time computation: the \texttt{thinking} variant of Qwen3-VL-235B reaches the same 19\% rotated accuracy as \texttt{instruct}. We also tested instruction tuning using six different styles, including two that hand the model an explicit procedure for rotating the scene or anchoring on a landmark. 

What is left is the resolution of the representation itself, and the distractor manipulation measures it. Moving the alternatives from 6.8\,m to 31.4\,m apart, on the same questions with the same target and the same answer positions, takes Gemini 3.8 Flash from 39\% to 85\% and GPT-5.6 Luna from 31\% to 55\%. A model holding no layout information could not gain 46 points from a change that only moves the distractors, and a model holding layout at human resolution would not need 30\,m of separation before it could use it. The allocentric map exists but its coarse.

\paragraph{Limitations.}
The scaling series are restricted to open-weight models, as frontier parameter counts remain undisclosed. In the landmark-count banks, the number of landmarks and the layout separation co-vary, meaning the axis measures layout separation rather than set size in isolation. Finally, as the stimuli are synthetic, exposure to similar rendered environments during pre-training could provide advantages. The human baseline is based on one participant which establishes that the task is solvable. More participants are currently being evaluated. 

\bibliographystyle{plainnat}
\bibliography{refs}

\newpage
\appendix
\begin{table}[t]
\centering
\small
\setlength{\tabcolsep}{6pt}
\begin{tabular}{@{}lccccc@{}}
\toprule
 & \includegraphics[width=0.122\textwidth]{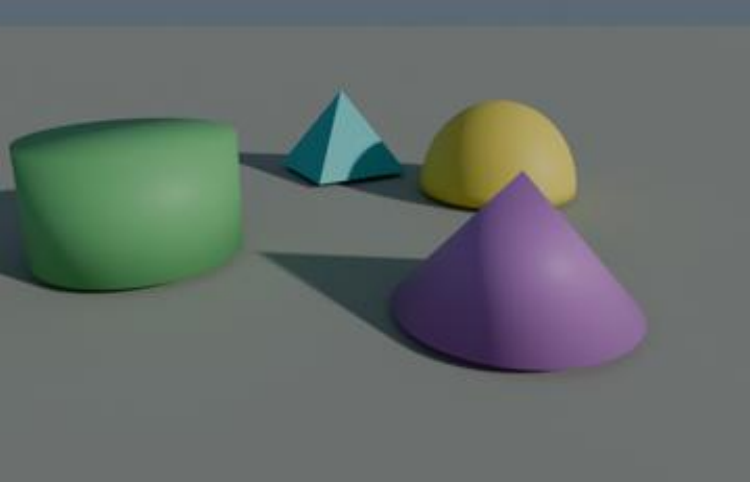} & \includegraphics[width=0.122\textwidth]{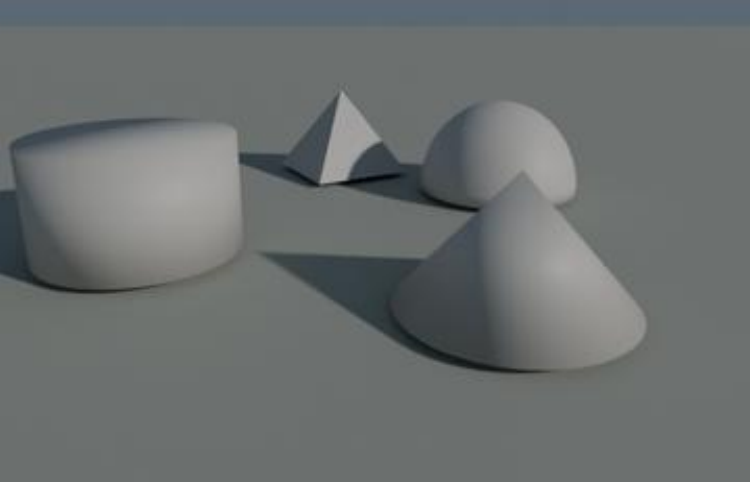} & \includegraphics[width=0.122\textwidth]{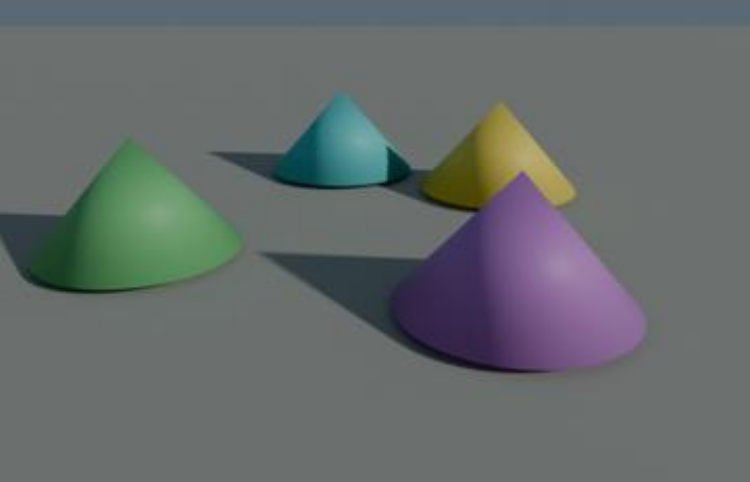} & \includegraphics[width=0.122\textwidth]{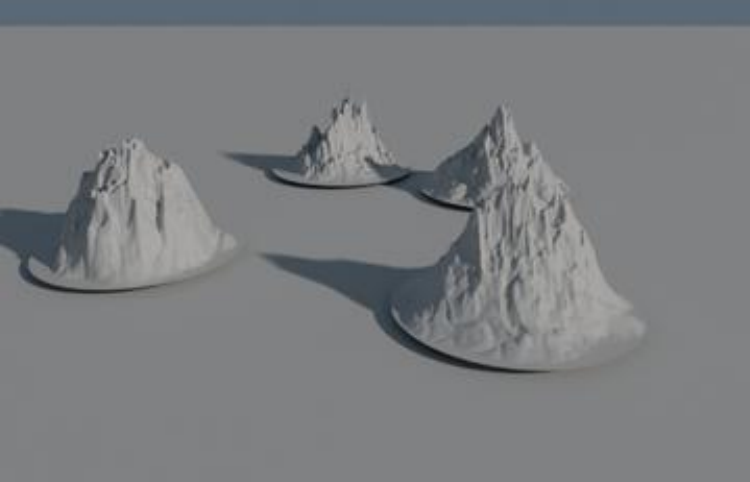} & \includegraphics[width=0.122\textwidth]{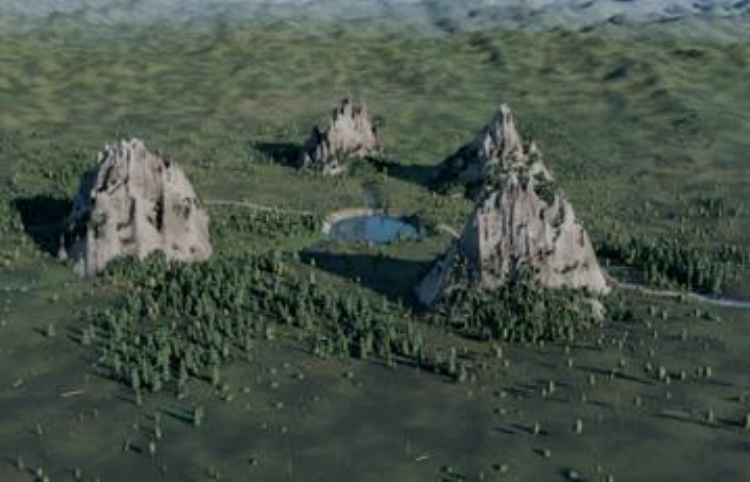} \\[1pt]
\textbf{Observer} & \textbf{c0} & \textbf{c1} & \textbf{c2} & \textbf{c3} & \textbf{c4} \\
\midrule
Human & \textbf{94\%} & \textbf{75\%} & \textbf{81\%} & \textbf{81\%} & \textbf{81\%} \\
\midrule
InternVL3.5-4B & 19\% & 6\% & 25\% & \textbf{0\%} & 12\% \\
InternVL3.5-2B & 12\% & \textbf{12\%} & 19\% & 25\% & 31\% \\
InternVL3.5-1B & \textbf{12\%} & 19\% & 25\% & 25\% & 19\% \\
InternVL3.5-38B & 19\% & \textbf{6\%} & 19\% & 19\% & 31\% \\
InternVL3.5-8B & 19\% & \textbf{6\%} & 19\% & 44\% & 19\% \\
InternVL3.5-30B-A3B & \textbf{6\%} & 19\% & 19\% & 31\% & 38\% \\
Qwen2.5-VL-32B & 12\% & \textbf{0\%} & 31\% & 6\% & 38\% \\
Qwen2.5-VL-3B & 38\% & \textbf{12\%} & 31\% & 38\% & 25\% \\
Qwen3-VL-235B (t) & 12\% & \textbf{12\%} & 19\% & 19\% & 31\% \\
InternVL3.5-14B & 25\% & \textbf{6\%} & 31\% & 38\% & 31\% \\
Qwen2.5-VL-72B & 19\% & 12\% & \textbf{12\%} & 31\% & 25\% \\
Qwen3-VL-235B (i) & 12\% & \textbf{6\%} & 19\% & 25\% & 31\% \\
InternVL3-38B & 25\% & 12\% & \textbf{12\%} & 38\% & 19\% \\
Qwen2.5-VL-7B & 50\% & 19\% & 25\% & \textbf{19\%} & 44\% \\
Gemini 3.8 Flash & 38\% & 38\% & 44\% & 44\% & \textbf{31\%} \\
GPT-5.6 Luna & 38\% & 19\% & 50\% & 31\% & \textbf{19\%} \\
\midrule
\textbf{All 16 models pooled} & \textbf{22\%} & \textbf{13\%} & \textbf{25\%} & \textbf{27\%} & \textbf{28\%} \\
\bottomrule
\vspace{0.1pt}
\end{tabular}
\caption{Rotated-trial accuracy by stimulus mode, 16 trials per cell, with one place shown in each mode above its column. c0 shape and colour, c1 shape only, c2 colour only, c3 bare peaks, c4 valley. Rows are ordered by overall accuracy. Bold marks the human row and, for each model, its weakest mode}
\label{tab:modes}
\end{table}

\begin{figure}[htbp]
\centering
\includegraphics[width=\textwidth]{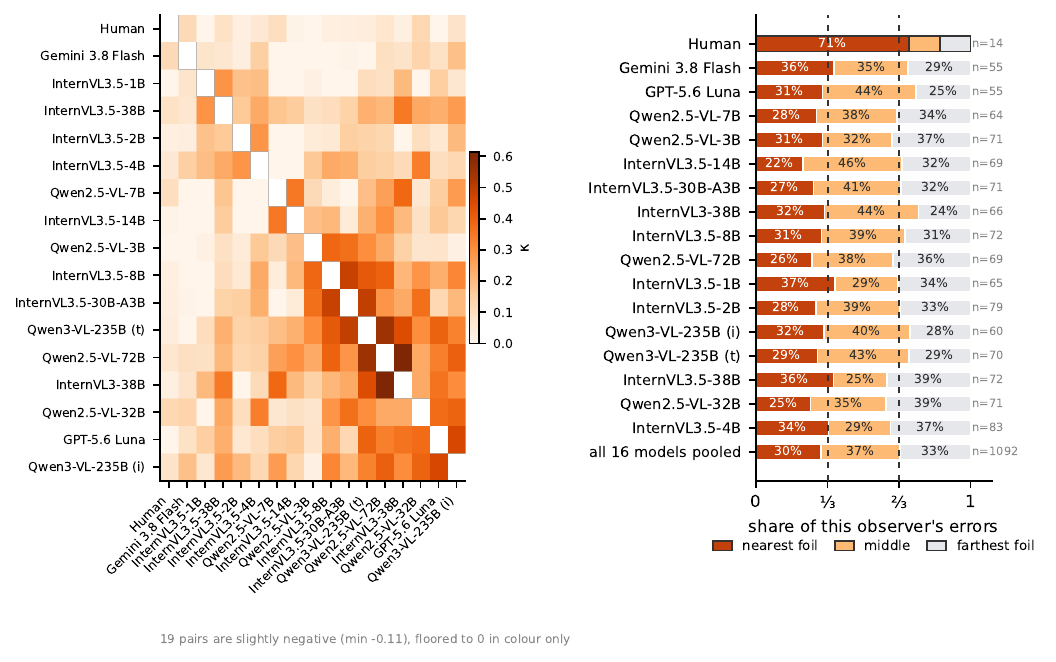}
\caption{\textbf{Left:} do a pair of observers agree on which trials are
solvable? Cohen's $\kappa$ on trial-level correctness; higher values mean a pair
succeeds and fails on the same trials. The human observer is in the first row,
with models ordered by mutual similarity. \textbf{Right:} when an observer is
wrong, which distractor does it pick? Distractors are ranked by layout distance
to the target, and the dashed lines mark thirds, which is what errors
independent of layout geometry would give.}
\label{fig:agree}
\end{figure}

\section{Additional results}

Table~\ref{tab:modes} splits the rotated trials by stimulus mode. Four of the five modes sit within each other's confidence intervals. The exception is c1, shape without colour, where the models answer 33 of 256 trials correctly (12.9\%, Wilson [9.3, 17.6]) against 25.5\% [22.9, 28.2] pooled over the other four (two-proportion $z=4.3$, $p=2\times10^{-5}$). It is the weakest cell for six of the sixteen models outright and joint-weakest for six more. Colour alone (c2) costs nothing by
comparison, at 25.0\%. It is consistent with colour carrying more identity information than shape in these renders, which would make c1 the mode with the least appearance to work from. It does not bear on the viewpoint result: every mode is far below the human, at every distractor separation.

\newpage
\subsection{The instructions, verbatim}
\label{app:prompts}
Every model saw one of these, followed by the study image and the four options.
The sweep in Figure~\ref{fig:landmarks} runs the first six on Qwen2.5-VL-32B;
every other run in the paper uses \texttt{cot}.

\paragraph{\texttt{cot}} chain of thought, viewpoint unstated.

\begin{footnotesize}
\begin{verbatim}
You are taking the Four Mountains Test of spatial allocentric perception.

Image 1 is the STUDY view of a landscape with four mountain peaks.
The subsequent 4 images are OPTION 1 to OPTION 4.
Exactly ONE option shows the EXACT SAME mountain landscape (the same four peaks in
  the same relative spatial arrangement). It may be viewed from the same direction
  as the study image or from a different one, and the lighting/weather may differ.
The other options show different mountain landscapes.

In 2-4 sentences, compare the 3D spatial layout of the peaks (e.g. relative
  positions such as in front, behind, left, right) between the study view and the
  options, accounting for any camera rotation. Avoid lengthy itemized lists.

State your final decision on the last line as:
Final Answer: Option X
\end{verbatim}
\end{footnotesize}

\paragraph{\texttt{cot + view}} chain of thought, viewpoint asserted.

\begin{footnotesize}
\begin{verbatim}
You are taking the Four Mountains Test of spatial allocentric perception.

Image 1 is the STUDY view of a landscape with four mountain peaks.
The subsequent 4 images are OPTION 1 to OPTION 4.
Exactly ONE option shows the EXACT SAME mountain landscape (the same four peaks in
  the same relative spatial arrangement), simply viewed from a different viewpoint
  and under different lighting/weather.
The other options show different mountain landscapes.

In 2-4 sentences, compare the 3D spatial layout of the peaks (e.g. relative
  positions such as in front, behind, left, right) between the study view and the
  options, accounting for camera rotation. Avoid lengthy itemized lists.

State your final decision on the last line as:
Final Answer: Option X
\end{verbatim}
\end{footnotesize}

\paragraph{\texttt{mental\_rotation}} instructed to mentally rotate.

\begin{footnotesize}
\begin{verbatim}
You are taking the Four Mountains Test of allocentric spatial perception.

Image 1 is the STUDY view of a landscape with four distinct mountain peaks.
The subsequent 4 images are OPTION 1 to OPTION 4.
Exactly ONE option shows the EXACT SAME mountain landscape (the same four peaks in
  the same relative geometric layout), viewed from a shifted camera viewpoint and
  under different weather/lighting.
The other options are distractors where the relative 3D spatial arrangement of the
  peaks has been altered.

To determine the matching scene, perform mental rotation:
1. Estimate the camera perspective shift (rotation angle) between the study image
  and candidate options.
2. Mentally rotate the four peaks to verify if topological handedness (e.g.
  clockwise/counterclockwise order of peaks, which peak is opposite or between
  others) matches the study scene.
3. Ignore superficial differences in sunlight, shadows, fog, and seasonal color.

In 2-4 sentences, explain your mental rotation reasoning, then conclude on the last
  line:
Final Answer: Option X
\end{verbatim}
\end{footnotesize}

\paragraph{\texttt{anchor}} instructed to pick an anchor landmark.

\begin{footnotesize}
\begin{verbatim}
You are taking the Four Mountains Test of allocentric spatial perception.

Image 1 is the STUDY view of a landscape with four distinct mountain peaks.
The subsequent 4 images are OPTION 1 to OPTION 4.
Exactly ONE option shows the EXACT SAME mountain landscape under a camera viewpoint
  rotation and weather change.
The remaining options are distractors.

Spatial Strategy:
1. Select the single most prominent or unique landmark peak as an anchor (origin).
2. Trace the relative bearings and distances of the other 3 peaks surrounding this
  anchor.
3. Identify which option preserves this exact 3D spatial configuration around the
  anchor under the new camera angle.

In 2-4 concise sentences, explain your reasoning and conclude on the last line:
Final Answer: Option X
\end{verbatim}
\end{footnotesize}

\paragraph{\texttt{birdseye}} instructed to imagine a plan view.

\begin{footnotesize}
\begin{verbatim}
You are taking the Four Mountains Test of allocentric spatial perception.

Image 1 is the STUDY view of a landscape with four mountain peaks.
The subsequent 4 images are OPTION 1 to OPTION 4.
Exactly ONE option shows the EXACT SAME mountain landscape viewed from a different
  camera viewpoint and lighting.
The other options are distractors with altered 3D mountain configurations.

Top-Down Cognitive Mapping Strategy:
1. Imagine looking down at the four peaks from directly above (a 2D bird's-eye map).
  Note their relative positions (which forms a triangle, which is isolated, which is
  tallest).
2. For each candidate option, determine where the camera would be standing on that
  same bird's-eye map.
3. Verify which option is geometrically consistent with the study scene's top-down
  layout under the new camera angle.

In 2-4 sentences, describe the bird's-eye spatial layout and conclude on the last
  line:
Final Answer: Option X
\end{verbatim}
\end{footnotesize}

\paragraph{\texttt{elimination}} instructed to eliminate alternatives.

\begin{footnotesize}
\begin{verbatim}
You are taking the Four Mountains Test of allocentric spatial perception.

Image 1 is the STUDY view of a landscape with four mountain peaks.
The subsequent 4 images are OPTION 1 to OPTION 4.
Exactly ONE option shows the EXACT SAME mountain landscape viewed from a different
  camera angle and lighting.
The other options are geometric distractors.

Falsification Strategy:
1. Inspect each candidate option one by one to find geometric contradictions with
  the study scene (e.g. impossible relative peak heights, wrong peak ordering, or
  missing ridges).
2. Eliminate the distractor options that cannot possibly match the study landscape
  under any viewpoint rotation.
3. Select the remaining single candidate that has no geometric contradictions.

Briefly eliminate the distractors and conclude on the last line:
Final Answer: Option X
\end{verbatim}
\end{footnotesize}

\paragraph{\texttt{neutral\_anyview}} the human arm, and the only wording a person saw.

\begin{footnotesize}
\begin{verbatim}
You will see a STUDY image of a place, then 4 options.

The place contains several landmarks. Exactly ONE option shows the SAME place as the
  study image. It may be photographed from the same direction as the study image or
  from a different one, and the lighting may differ.
The other options show different places, each containing the same landmarks arranged
  differently, photographed from the same direction as the correct option.

Which option shows the same place as the study image?
Answer on the last line as:
Final Answer: Option X
\end{verbatim}
\end{footnotesize}

\end{document}